\documentclass[11pt]{article} % For LaTeX2e
\usepackage{nips15submit_e,times}
\usepackage[acronym]{glossaries}
\usepackage{color}
\usepackage{cite} 
\usepackage{amsmath,amsfonts,amssymb}
\usepackage{amsthm}
\usepackage{graphicx} 
\usepackage{cancel}
\usepackage{verbatim}
\usepackage{url}
\usepackage{xcolor}
\usepackage{multirow}
\usepackage{caption}
\usepackage{subcaption}
\usepackage{algpseudocode}
\usepackage{algorithm}
\usepackage{tikz}
\usepackage{booktabs,siunitx}
\newcommand{\boldentry}[1]{%
  \multicolumn{1}{S[table-format=-1.2, 
                    table-figures-uncertainty=1,
                    mode=text,
                    text-rm=\fontseries{b}\selectfont
                   ]}{#1}}
\usetikzlibrary{arrows}

\algnewcommand{\Initialize}[1]{%
	\State \textbf{Initialize:}
	\Statex \hspace*{\algorithmicindent}\parbox[t]{.8\linewidth}{\raggedright #1}
}

\graphicspath{{./figures/}}

\newglossaryentry{sde}{%
  name={sde},%
  description={stochastic differential equation},%
  first={stochastic differential equation (SDE)},%
  firstplural={stochastic differential equations (SDEs)},%
  text={SDE},%
  plural={SDEs}
}

\newglossaryentry{rmse}{%
	name={rmse},%
	description={root mean squared error},%
	first={root mean squared errors (RMSE)},%
	firstplural={root mean squared errors (RMSEs)},%
	text={RMSE},%
	plural={RMSEs}
}

\newglossaryentry{nll}{%
	name={nll},%
	description={negative log loss},%
	first={negative log loss (NLL)},%
	firstplural={negative log loss (NLLs)},%
	text={NLL},%
	plural={NLLs}
}

\newglossaryentry{gp}
{% 
  name={GP},% 
  description={Gaussian Process},% 
  first={Gaussian Process (GP)},% 
  firstplural={Gaussian Processes (GPs)},%
  text={GP}%
}

\newglossaryentry{dgp}
{% 
  name={DGP},% 
  description={deep Gaussian Process},% 
  first={deep Gaussian Process (DGP)},% 
  firstplural={deep Gaussian Processes (DGPs)},%
  text={DGP}%
}

\newglossaryentry{mlp}
{% 
	name={MLP},% 
	description={multi layer perceptron},% 
	first={multi layer perceptron (MLP)},% 
	firstplural={multi layer perceptrons (MLPs)},%
	text={MLP}%
}

\newglossaryentry{sep}
{% 
  name={SEP},% 
  description={stochastic expectation propagation},% 
  first={Stochastic Expectation Propagation (SEP)},% 
  text={SEP}%
}

\newglossaryentry{ep}
{% 
  name={EP},% 
  description={expectation propagation},% 
  first={Expectation Propagation (EP)},% 
  text={EP}%
}

\newglossaryentry{gplvm}
{% 
  name={GPLVM},% 
  description={Gaussian Process Latent Variable Model},% 
  first={Gaussian Process Latent Variable Model (GPLVM)},% 
  firstplural={Gaussian Process Latent Variable Models (GPLVMs)},%
  text={GPLVM}%
}

\newglossaryentry{gpssm}
{% 
  name={GPSSM},% 
  description={Gaussian Process State Space Model},% 
  first={Gaussian Process State Space Model (GPSSM)},% 
  firstplural={Gaussian Process State Space Models (GPSSMs)},%
  text={GPSSM}%
}

\newglossaryentry{fitc}%
{%
  name={FITC},%
  description={Fully Independent Training Conditional},%
  first={Fully Independent Training Conditional (FITC)},%
  text={FITC}
}

\newcommand{\norm}{\mathcal{N}}
\newcommand{\xvec}{\mathbf{x}}
\newcommand{\zvec}{\mathbf{z}}

\newcommand{\uvec}{\mathbf{u}}
\newcommand{\yvec}{\mathbf{y}}

\newcommand{\hvec}{\mathbf{h}}

\newcommand{\Xvec}{\mathbf{X}}

\newcommand{\Hvec}{\mathbf{H}}

\newcommand{\zero}{\mathbf{0}}

\newcommand{\dd}{\mathrm{d}}

\DeclareMathOperator{\tr}{tr}

\allowdisplaybreaks[4]

\newcommand*{\mathcolor}{}
\def\mathcolor#1#{\mathcoloraux{#1}}
\newcommand*{\mathcoloraux}[3]{%
  \protect\leavevmode
  \begingroup
    \color#1{#2}#3%
  \endgroup
}

\makeatletter
\newcommand{\mypm}{\mathbin{\mathpalette\@mypm\relax}}
\newcommand{\@mypm}[2]{\ooalign{%
		\raisebox{.1\height}{$#1+$}\cr
		\smash{\raisebox{-.6\height}{$#1-$}}\cr}}
\makeatother

\usepackage{cleveref}
\usepackage[backgroundcolor=white,linecolor=red,bordercolor=white,textsize=tiny,textwidth=10mm]{todonotes}

\usepackage{titlesec}
\titlespacing{\section}{3pt}{0.9ex}{0.7ex}
\titlespacing{\subsection}{3pt}{0.5ex}{0ex}
\titlespacing{\subsubsection}{3pt}{0.2ex}{0ex}

\nipsfinalcopy

\title{Training Deep Gaussian Processes using Stochastic Expectation Propagation and Probabilistic Backpropagation}

\author{
Thang D. Bui \\
University of Cambridge\\
\texttt{tdb40@cam.ac.uk} \\
\And
Jos\'e Miguel Hern\'andez-Lobato \\
Harvard University \\
\texttt{jmhl@seas.harvard.edu} \\
\AND
Yingzhen Li \\
University of Cambridge \\
\texttt{yl494@cam.ac.uk} \\
\And
Daniel Hern\'andez-Lobato \\
Universidad Aut\'onoma de Madrid \\
\texttt{daniel.hernandez@uam.es} \\
\And
Richard E. Turner \\
University of Cambridge \\
\texttt{ret26@cam.ac.uk} \\
}

\begin{document}
\maketitle
\begin{abstract}
Deep Gaussian processes (DGPs) are multi-layer hierarchical generalisations of Gaussian processes (GPs) and are formally equivalent to neural networks with multiple, infinitely wide hidden layers. DGPs are probabilistic and non-parametric and as such are arguably more flexible, have a greater capacity to generalise, and provide better calibrated uncertainty estimates than alternative deep models. The focus of this paper is scalable approximate Bayesian learning of these networks. The paper develops a novel and efficient extension of probabilistic backpropagation, a state-of-the-art method for training Bayesian neural networks, that can be used to train DGPs. The new method leverages a recently proposed method for scaling Expectation Propagation, called stochastic Expectation Propagation. The method is able to automatically discover useful input warping, expansion or compression, and it is therefore is a flexible form of Bayesian kernel design. We demonstrate the success of the new method for supervised learning on several real-world datasets, showing that it typically outperforms GP regression and is never much worse.
\end{abstract}
%%%%%%%%%%%%%%%%%%%%%%%%%%%%%%%%%%%%%%%%%
\section{Introduction}
\glspl{gp} are powerful nonparametric distributions over continuous functions which can be used for both supervised and unsupervised learning problems \cite{rasmussen2005gpml}. In this article, we study a multi-layer hierarchical generalisation of \glspl{gp} or \glspl{dgp} for supervised learning tasks. A \gls{gp} is equivalent to an infinitely wide neural network with single hidden layer and similarly a \gls{dgp} is a multi-layer neural network with multiple infinitely wide hidden layers \cite{damianou-lawrence:2013a}. The mapping between layers in this type of network is parameterised by a \gls{gp}, and as a result \glspl{dgp} are arguably more flexible, have a greater capacity to generalise, and are able to provide better calibrated predictive uncertainty estimates than standard multi-layer models \cite{andreasthesis}.

More formally, suppose we have a training set comprising $N$ $D$-dimensional input vectors $\{\xvec_n\}_{n=1}^{N}$ and corresponding real valued scalar observations $\{y_n\}_{n=1}^{N}$. The probabilistic representation of a \gls{dgp} comprising of $L$ layers can be written as follows,
\small
\begin{align}
	p(f_{l}|\theta_l) &= \mathcal{GP}(f_{l};\zero,\mathbf{K}_{l}), \;\; {l=1,\cdots,L}\\
	p(\hvec_{l}|f_{l},\hvec_{l-1},\sigma_l^2) &=\prod_{n}\mathcal{N}(h_{l,n};f_{l}(h_{l-1,n}),\sigma_{l}^2), \;\; h_{1,n} = \xvec_n\\
	p(\yvec|f_{L},\Xvec_{L-1},\sigma_L^2) &=\prod_{n}\mathcal{N}(y_{L,n};f_{L}(h_{L-1,n}),\sigma_{L}^2)
\end{align}
\normalsize
Here hidden layers are denoted $h_{l,n}$ and the mapping function between the layers, $f_{l}$, is drawn from a \gls{gp}. A Gaussian (regression) likelihood is used in this work, a different likelihood can easily be accommodated.\footnote{Hidden variables in the intermediate layers can and will generally have multiple dimensions but we have omitted this here to lighten the notation.} 

The \gls{dgp} collapses back to a standard \gls{gp} when $L=1$  (when there are no hidden layers) or when only one of the functions is non-linear. The addition of non-linear hidden layers can potentially overcome practical limitations of {\it shallow} \glspl{gp}. First, modelling real-world complex datasets often requires rich, hand-designed covariance functions. \glspl{dgp} can perform input warping or dimensionality compression or expansion, and hence automatically learn to construct a kernel that works well for the data at hand. Second, the functional mapping from inputs to outputs specified by a \gls{dgp} is non-Gaussian which is a more general and flexible modelling choice. Third, \glspl{dgp} can repair damage done by sparse approximations to the representational power of each \gls{gp} layer. For example, inducing point based approximation methods for \glspl{gp} trade model complexity for a lower computational complexity of $\mathcal{O}(LNM^2)$ where $L$ is the number of layers, $N$ is the number of datapoints and $M$ is the the number of inducing points. This complexity scales quadratically in $M$ whereas the dependence on the number of layers $L$ is only linear. Therefore, it can be cheaper to increase the representation power of the model by adding extra layers rather than adding more inducing points.

The focus of this paper is Bayesian learning of \glspl{dgp}, which involves inferring the posterior over the mappings between layers and hyperparameter tuning using the marginal likelihood. Unfortunately, exact Bayesian learning in this model is analytically intractable and as such approximate inference is needed. Recent work in this frontier largely focussed on variational free-energy approaches \cite{hensman+lawrence:2014}. We introduce an alternative approximation scheme based on three approximations. First, in order to side step the cubic computational cost of GPs we leverage a well-known inducing point sparse approximation \cite{snelson+ghahramani:2006, quinonero+rasmussen:2005}. Second, expectation propagation is used to approximate the analytically intractable posterior over the inducing points. Here we utilise stochastic expectation propagation (SEP) that prevents the memory overhead from increasing with the number of datapoints \cite{li+etal:2015}. Third, the SEP moment computation is itself analytically intractable and requires one final approximation. For this we use the probabilistic backpropagation approximation \cite{hernandez+adams:2015}. The proposed enables the advantages of the \gls{dgp} model to be realised through a computationally efficient, scalable and easy to implement algorithm.
%%%%%%%%%%%%%%%%%%%%%%%%%%%%%%%%%%%%%%%%%

\section{The Fully Independent Training Conditional approximation\label{sec:fitc}}
The computational complexity of full \gls{gp} models scales cubically with the number of training instances, making it intractable in practice. Sparse approximation techniques are therefore often resorted to. They can be coarsely put into two classes: ones that explicitly sparsify and create a parametric representation that approximates the original model, and ones that retain the original nonparametric properties and perform sparse approximation to the exact posterior. The method we describe and use here, \gls{fitc}, falls into the first category. The \gls{fitc} approximation is formed by considering a small set of function values $\uvec$ in the infinite dimensional vector $f$ and assuming conditional independence between the remaining values given the set $\uvec$ \cite{snelson+ghahramani:2006,quinonero+rasmussen:2005}. This set is often called inducing points or pseudo datapoints and their input locations can be chosen by optimising the approximate marginal likelihood so that the approximate model is closer to the original model. The resulting model can be written as follows,
\begin{align}
	p(\uvec_{l}|\theta_l) &= \norm(\uvec_{l};\zero,\mathbf{K}_{\zvec_{l-1},\zvec_{l-1}}), \;\; {l=1,\cdots,L}\\
	p(\hvec_{l}|\uvec_{l},\hvec_{l-1},\sigma_l^2) &=\prod_{n}\mathcal{N}(h_{l,n}; \mathbf{C}_{n,l} \uvec_{l}, \mathbf{R}_{n,l}),\\
	p(\yvec|\uvec_{L},\Hvec_{L-1},\sigma_L^2) &=\prod_{n}\mathcal{N}(y_{n}; \mathbf{C}_{n,L} \uvec_{L}, \mathbf{R}_{n,L}).
\end{align}
where $\mathbf{C}_{n,l} = \mathbf{K}_{\hvec_{l-1,n},\zvec_{l-1}} \mathbf{K}_{\zvec_{l-1},\zvec_{l-1}}^{-1}$ and $\mathbf{R}_{n,l} = \mathbf{K}_{\hvec_{l-1,n},\hvec_{l-1,n}} - \mathbf{K}_{\hvec{l-1,n},\zvec_{l-1}} \mathbf{K}_{\zvec_{l-1},\zvec_{l-1}}^{-1} \mathbf{K}_{\zvec_{l-1},\hvec_{l-1,n}} + \sigma_l^2$.  The \gls{fitc} approximation creates a parametric model, but one which is cleverly structured so that the induced non-stationary noise captures the uncertainty introduced from the sparsification. 
%%%%%%%%%%%%%%%%%%%%%%%%%%%%%%%%%%%%%%%%%

%%%%%%%%%%%%%%%%%%%%%%%%%%%%%%%%%%%%%%%%%
\section{Stochastic expectation propagation for deep, sparse Gaussian processes}
Having specified a probabilistic model for data using a deep sparse Gaussian processes we now consider inference for the inducing outputs $\uvec$ and learning of the inducing inputs $\{\zvec_l\}_{l=1}^L$ and hyperparameters $\{ \theta \}_{l=1}^L$. The posterior distribution over the inducing points can be written as $p(\uvec|\Xvec,\yvec) \propto p(\uvec) \prod_n p(y_n|\uvec,\Xvec_n)$. This quantity can then be used for prediction of output given a test input, $p(y^*|\xvec^*,\Xvec,\yvec) = \int \dd\uvec p(\uvec|\Xvec,\yvec) p(y^*|\uvec,\xvec^*)$. However, the posterior of $\uvec$ is not analytically tractable when there is more than one \gls{gp} layer in the model. As such, approximate inference is needed; here we use \gls{sep}, a recently proposed modification to \gls{ep} \cite{li+etal:2015}.

In \gls{sep}, the posterior $p(\uvec|\Xvec, \yvec)$ is approximated by $q(\uvec) \propto p(\uvec) g(\uvec)^{N}$, where the factor $g(\uvec)$ could be thought of as an {\it average} data factor that captures the average effect of a likelihood term on the posterior. The form chosen, though seems limited as first, in practice performs almost as well as \gls{ep} in which there is a factor $g_n(\uvec)$ per datapoint, while significant reducing EP's memory footprint \cite{li+etal:2015}. Specifically for our model, the memory complexity of \gls{ep} is $\mathcal{O}(NM^2)$ as we need to store the mean and covariance matrix for each data factor; in contrast, such requirement for \gls{sep} is only $\mathcal{O}(M^2)$ regardless of the number of training points. 

The \gls{sep} procedure involves looping through the dataset multiple times and performing the following steps: 1.\ remove $g(\uvec)$ from the approximate posterior to form the cavity distribution, 2.\ incorporate a likelihood term into the cavity to form the tilted distribution, 3.\ moment match the approximate posterior to this distribution, and in addition to \gls{ep}, 4.\ perform a small update to the {\it average} factor $g(\uvec)$. We choose a Gaussian form for both $q(\uvec)$  and $g(\uvec)$, and as a result steps 1 and 4 are analytically tractable. We will discuss how to deal with the intermediate steps given a training datapoint $(\xvec, y)$  in the next section.

\section{Probabilistic backpropagation for deep, sparse Gaussian processes}
The moment matching step in SEP is analytically intractable as it involves propagating a Gaussian distribution through a \gls{dgp} and computing the moments of the resulting complex distribution. However, for certain choices of covariance functions $\{ \mathbf{K}_{l} \}_{l=1}^L$, it is possible to use an efficient and accurate approximation which propagates a Gaussian through the first layer of the network and projects this non-Gaussian distribution back to a moment matched Gaussian before propagating through the next layer and repeating the same steps. This scheme is a central part of the probabilistic backpropagation algorithm that has been applied to standard neural networks \cite{hernandez+adams:2015}.

In more detail, let $q^{\setminus 1}(\uvec) = \norm(\uvec; \mathbf{m}^{\setminus 1}, \mathbf{V}^{\setminus 1})$ be the cavity distribution, the difficult steps above are equivalent to the following updates to the mean and covariance of the approximate posterior:
\small
\begin{align}
\mathbf{m} &= \mathbf{m}^{\setminus 1} + \mathbf{V}^{\setminus 1} \frac{\dd\log \mathcal{Z}}{\dd\mathbf{m}^{\setminus 1}}, &
\mathbf{V} &= \mathbf{V}^{\setminus 1} - \mathbf{V}^{\setminus 1} \left[\frac{\dd\log \mathcal{Z}}{\dd\mathbf{m}^{\setminus 1}} \left(\frac{\dd\log \mathcal{Z}}{\dd\mathbf{m}^{\setminus 1}}\right)^{\intercal} -2\frac{\dd\log \mathcal{Z}}{\dd\mathbf{V}^{\setminus 1}}\right]\mathbf{V}^{\setminus 1},
\end{align}
\normalsize
where $\mathcal{Z} = \int \dd \uvec p(y|\xvec, \uvec) q^{\setminus 1}(\uvec)$ \cite{minka2001family}. The inference scheme therefore reduces to evaluating the normalising constant $\mathcal{Z}$ and its gradient. By reintroducing the hidden variables in the middle layers, we perform Gaussian approximation to $\mathcal{Z}$ in a sequential fashion, taking a two \gls{gp} layers case as an example:
\small
\begin{align}
\mathcal{Z} 
&= \int \dd \uvec p(y|\xvec, \uvec) q^{\setminus 1}(\uvec) = \int \dd h_1 \dd \uvec_2 p(y|h_1, \uvec_2) q^{\setminus 1}(\uvec_2) \int \dd \uvec_1 p(h_1|\xvec, \uvec_1) q^{\setminus 1}(\uvec_1)
\end{align}
\normalsize
We can exactly marginalise out the inducing points for each \gls{gp} layer leading to $\mathcal{Z} = \int \dd h_1 \dd \uvec_2 q(y|h_1) q(h_1)$ where $q(h_1) = \norm(h_1; m_1, v_1) $, $q(y|h_1) = \norm(y|h_1; m_{2|h_1}, v_{2|h_1})$ and
\small
\begin{align}
&m_1 = \mathbf{K}_{\xvec,\zvec_1} \mathbf{K}_{\zvec_1,\zvec_1}^{-1} \mathbf{m}_1^{\setminus 1}, \quad v_1 = \sigma_1^2 + K_{\xvec,\xvec} - \mathbf{K}_{\xvec,\zvec_1} \mathbf{K}_{\zvec_1,\zvec_1}^{-1} \mathbf{K}_{\zvec_1,\xvec} + \mathbf{K}_{\xvec,\zvec_1} \mathbf{K}_{\zvec_1,\zvec_1}^{-1} \mathbf{V}_1^{\setminus 1} \mathbf{K}_{\zvec_1,\zvec_1}^{-1} \mathbf{K}_{\zvec_1,\xvec} \nonumber\\
&m_{2|h_1} = \mathbf{K}_{h_1,\zvec_2} \mathbf{K}_{\zvec_2,\zvec_2}^{-1} \mathbf{m}_2^{\setminus 1}, \quad v_{2|h_1} = \sigma_2^2 + K_{h_1,h_1} - \mathbf{K}_{h_1,\zvec_2} \mathbf{K}_{\zvec_2,\zvec_2}^{-1} \mathbf{K}_{\zvec_2, h_1} + \mathbf{K}_{h_1,\zvec_2} \mathbf{K}_{\zvec_2,\zvec_2}^{-1} \mathbf{V}_1^{\setminus 1} \mathbf{K}_{\zvec_2,\zvec_2}^{-1} \mathbf{K}_{\zvec_2,h_1} \nonumber
\end{align}
\normalsize
Following \cite{GirRasQuiMur03}, we can approximate the difficult integral in the equation above by a Gaussian $\mathcal{Z} \approx \norm(y|m_2, v_2)$ where the mean and variance take the following form,
\small
\begin{align}
m_{2} &= \mathrm{E}_{q(h_1)} [m_{2|h_1}] = \mathrm{E}_{q(h_1)} [\mathbf{K}_{h_1,\zvec_2}] \mathbf{K}_{\zvec_2,\zvec_2}^{-1} \mathbf{m}_2^{\setminus 1}\\
v_{2} &= \mathrm{E}_{q(h_1)} [v_{2|h_1}]  + \mathrm{var}_{q(h_1)}[m_{2|h_1}] \\
		&= \sigma_2^2 + \mathrm{E}_{q(h_1)} [K_{h_1,h_1}] + \tr \left( \mathbf{B} \mathrm{E}_{q(h_1)} [\mathbf{K}_{\zvec_2,h_1} \mathbf{K}_{h_1,\zvec_2} ] \right) - m_{2}^2
\end{align}
\normalsize
where \small$\mathbf{B} = \mathbf{K}_{\zvec_2,\zvec_2}^{-1} ( \mathbf{V}_2^{\setminus 1} + \mathbf{m}_2^{\setminus 1} \mathbf{m}_2^{\setminus 1, \mathrm{T}} ) \mathbf{K}_{\zvec_2,\zvec_2}^{-1} - \mathbf{K}_{\zvec_2,\zvec_2}^{-1}$\normalsize. The equations above require the expectations of the kernel matrix under a Gaussian distribution over the inputs, which are analytically tractable for widely used kernels such as exponentiated quadratic, linear or a more general class of spectral mixture kernels \cite{wilson+adams:2013}. Importantly, the computation graph of the approximation to $\log \mathcal{Z}$ and its gradient can be easily programmed using symbolic packages such as Theano \cite{Bastien-Theano-2012}.
%%%%%%%%%%%%%%%%%%%%%%%%%%%%%%%%%%%%%%%%%

%%%%%%%%%%%%%%%%%%%%%%%%%%%%%%%%%%%%%%%%%
\section{Stochastic optimisation of hyperparameters and inducing point locations}
We complete the main text by discussing the optimisation of model hyperparameters and inducing point locations. Given the approximation to the posterior obtained by using \gls{sep} as described above, one can also obtain the approximate marginal likelihood and its gradients. As a result, parameter training now involves iterating between running \gls{sep} and updating the hypeparameters based on these gradients. However, this procedure can only made efficient and scalable by following two observations discussed in \cite{daniel+miguel:2015}, which include 1.\ we do not need to wait for (S)EP to converge before making an update to the parameters, and 2.\ the gradients involve a sum across the whole training set, enabling fast optimisation using stochastic gradients computed on minibatches of datapoints.
%%%%%%%%%%%%%%%%%%%%%%%%%%%%%%%%%%%%%%%%%
%\todo{mention speed here somewhere and compare/contract to variational if there is space}

%%%%%%%%%%%%%%%%%%%%%%%%%%%%%%%%%%%%%%%%%
\section{Experimental results}
We test our approximation method on several \gls{dgp} architectures for a regression task on several real-world datasets. We obtain 20 random splits of each dataset, 90\% for training and 10\% for testing and report the average results and their standard deviations in table \ref{tab:res}. The prediction errors are evaluated using two metrics: root mean squared error (RMSE) and mean log loss (MLL). We use an exponentiated quadratic kernel with ARD lengthscales. The lengthscales and inducing points of the first \gls{gp} layer are sensibly initialised based on the median distance between datapoints in the input space and the k-means cluster centers respectively. We use long lengthscales and initial inducing points between $[-1, 1]$ for the higher layers to force them to start up with an identity mapping. For all results reported here, we use Adam \cite{kingma+ba:2015} with minibatch size of 50 datapoints and run the optimiser for 4000 iterations. The learning rate is selected by optimising the predictive errors on a small subset of training points using Bayesian optimisation \cite{snoek2012practical}. We experiment with two one-hidden-layer \gls{dgp} networks with hidden variables of one and two dimensions, 50 inducing points per layer and denote them as [DGP, 1, 50] \footnote{This is the same as Bayesian warped GPs, that is the one dimensional output of the first layer is warped through a \gls{gp} to form the prediction/output.} and [DGP, 2, 50] respectively. We compare them against sparse \gls{gp} regression with the same number of inducing points [GP, 50]. The results in the table below show that overall \gls{dgp} with two dimensional hidden variables perform as well or better than sparse \gls{gp}, and almost always better than the architecture with one dimensional hidden variables. Taking the Boston Housing dataset as an example, the mean test log likelihood using [DGP, 2, 50] is -2.12, which is, to the best of our knowledge, better than state-of-the-art results which were obtained using Bayesian neural networks with probabilistic backpropagation: -2.57 \cite{hernandez+adams:2015}, dropout: -2.46 \cite{yarin+zoubin:2015} or SGLD: -2.31 \cite{koretal:2015}.
\begin{table}[!ht]
\begin{center}
\scriptsize
\begin{tabular}{l 
c
c
S[table-format=-1.2, table-figures-uncertainty=1]
S[table-format=-1.2, table-figures-uncertainty=1]
S[table-format=-1.2, table-figures-uncertainty=1]
S[table-format=-1.2, table-figures-uncertainty=1]
S[table-format=-1.2, table-figures-uncertainty=1]
S[table-format=-1.2, table-figures-uncertainty=1]
}
 \toprule
 & & &  \multicolumn{3}{c}{RMSE} & \multicolumn{3}{c}{MLL} \\
 {Dataset} & {N} & {D} & {GP, 50} & {DGP, 1, 50} & {DGP, 2, 50} & {GP, 50} & {DGP, 1, 50} & {DGP, 2, 50} \\
 \midrule
boston  &    506    &    13     &    3.09 \pm 0.63  &    2.85 \pm 0.65  &    \boldentry{2.47 \pm 0.49}  &    -2.26 \pm 0.31     &    -2.30 \pm 0.53     &    \boldentry{-2.12 \pm 0.37}\\
concrete    &    1030   &    8  &    5.24 \pm 0.55  &    5.91 \pm 1.65  &    \boldentry{5.21 \pm 0.90}  &    -2.97 \pm 0.10     &    -3.07 \pm 0.14     &    \boldentry{-2.70 \pm 0.35}\\
energy 1    &    768    &    8  &    0.50 \pm 0.10  &    0.77 \pm 0.59  &    \boldentry{0.48 \pm 0.05}  &    -0.26 \pm 0.13     &    -0.39 \pm 0.37     &    \boldentry{-0.20 \pm 0.14}\\
energy 2    &    768    &    8  &    1.60 \pm 0.15  &    1.78 \pm 0.43  &    \boldentry{1.37 \pm 0.23}  &    -1.05 \pm 0.28     &    -1.14 \pm 0.32     &    \boldentry{-0.76 \pm 0.15}\\
kin8nm  &    8192   &    8  &    0.04 \pm 0.00  &    0.07 \pm 0.04  &    \boldentry{0.02 \pm 0.00}  &    2.02 \pm 0.06  &    1.71 \pm 0.35  &    \boldentry{2.48 \pm 0.03}\\
naval 1     &    11934  &    16     &    0.02 \pm 0.01  &    \boldentry{0.00 \pm 0.00}  &    0.00 \pm 0.00  &    2.64 \pm 1.14  &    \boldentry{5.14 \pm 0.37}  &    5.02 \pm 0.59\\
naval 2     &    11934  &    16     &    0.01 \pm 0.00  &    0.00 \pm 0.00  &    \boldentry{0.00 \pm 0.00}  &    3.52 \pm 0.02  &    4.67 \pm 0.68  &    \boldentry{5.24 \pm 0.48}\\
power   &    9568   &    4  &    3.19 \pm 0.18  &    3.35 \pm 0.20  &    \boldentry{2.95 \pm 0.30} &    -2.53 \pm 0.03     &    -2.61 \pm 0.05     &    \boldentry{-2.38 \pm 0.13}\\
red wine    &    1588   &    11     &    \boldentry{0.48 \pm 0.06}  &    0.62 \pm 0.05  &    0.54 \pm 0.11  &    -0.06 \pm 0.15     &    -0.10 \pm 0.64     &    \boldentry{0.29 \pm 0.65}\\
white wine  &    4898   &    11     &    0.37 \pm 0.04  &    0.49 \pm 0.09  &    \boldentry{0.34 \pm 0.07}  &    0.01 \pm 0.11  &    -0.17 \pm 0.36     &    \boldentry{0.66 \pm 0.31}\\
creep   &    2066   &    31     &    95.87 \pm 18.03    &    74.86 \pm 13.66    &    \boldentry{70.58 \pm 15.55}    &    -5.85 \pm 0.35     &    -5.45 \pm 0.20     &    \boldentry{-5.28 \pm 0.29}\\
\bottomrule
\end{tabular}
\end{center}
\caption{Predictive errors using \glspl{dgp} and \glspl{gp} for regression on several UCI datasets\label{tab:res}}
\end{table}

In addition, we also vary the number of inducing points per layer for the above networks and trace out the speed-accuracy frontier. Preliminary results indicates that \glspl{dgp} is very efficient using our inference technique and with a small number of inducing points, can obtain a predictive performance that would require many more inducing points in a shallower architecture.
%%%%%%%%%%%%%%%%%%%%%%%%%%%%%%%%%%%%%%%%%

%%%%%%%%%%%%%%%%%%%%%%%%%%%%%%%%%%%%%%%%%
\section{Conclusion}
We have proposed a novel approximation scheme for deep Gaussian processes for supervised learning. Our method extends probabilistic backpropagation for Bayesian neural networks, combines it with an inducing point based sparse \gls{gp} approximation and a recently proposed method for scalable approximate Bayesian inference, Stochastic Expectation Propagation. We systematically evaluate our approach on several regression datasets and the initial experimental results demonstrate the validity of our method and the effectiveness of \glspl{dgp} compared to \glspl{gp}. Our method is fast, easy to implement and promisingly, gives state-of-the-art performance in various regression tasks.

Current work includes performing experiments on large scale datasets, comparing our method to the variational approach presented in \cite{hensman+lawrence:2014}, extension to classification and unsupervised learning, and understanding the effect of the network architectures on prediction quality.
%%%%%%%%%%%%%%%%%%%%%%%%%%%%%%%%%%%%%%%%%

%%%%%%%%%%%%%%%%%%%%%%%%%%%%%%%%%%%%%%%%%
\subsubsection*{Acknowledgements}
TB thanks Google for funding his European Doctoral Fellowship.
JMHL acknowledges support from the Rafael del Pino Foundation.
DHL and JMHL acknowledge support from Plan Nacional I+D+i, Grant TIN2013-42351-P, and from CAM, Grant S2013/ICE-2845 CASI-CAM-CM.
YL thanks the Schlumberger Foundation for her Faculty for the Future PhD fellowship. 
RET thanks EPSRC grants EP/G050821/1 and EP/L000776/1. 
%%%%%%%%%%%%%%%%%%%%%%%%%%%%%%%%%%%%%%%%%%%%%%%%

%%%%%%%%%%%%%%%%%%%%%%%%%%%%%%%%%%%%%%%%%
\subsubsection*{References}
{\small
\renewcommand\refname{\vskip -1cm}
\bibliography{refs}
\bibliographystyle{ieeetr}
}
%%%%%%%%%%%%%%%%%%%%%%%%%%%%%%%%%%%%%%%%%%%%%%%%

\newpage

\end{document}